\documentclass[runningheads]{llncs}

\usepackage[T1]{fontenc}
\usepackage{graphicx}
\usepackage{verbatim}
\usepackage{amsmath}
\usepackage{amssymb}
\usepackage{booktabs}
\usepackage{multirow}
\usepackage{siunitx}
\usepackage{array}
\usepackage[table,xcdraw]{xcolor}
\usepackage{hyperref}
\usepackage{marvosym}
\usepackage{bm}
\usepackage{color}
\usepackage{tabularx}

\newcommand{\hl}[1]{\cellcolor[HTML]{D6EAF8}#1}
 
\DeclareMathOperator*{\argmax}{arg\,max}

\begin{document}

\title{Missing Modality-Aware Calibration for Trustworthy Brain Tumor Segmentation}

\titlerunning{MMA-LTS}
\author{Sol Lee \and
Hyunji Kim \and
Sungrae Hong \and
Donghee Han \and
Mun Yi\thanks{Corresponding author.}}
\authorrunning{S. Lee et al.}
\titlerunning{MMA-LTS}

\institute{
Korea Advanced Institute of Science and Technology, Daejeon, South Korea\\
\texttt{\{leesol4553, hyunjikim, sr5043, handonghee, munyi\}@kaist.ac.kr}}

\maketitle

\begin{abstract}
Multimodal brain tumor segmentation typically leverages multiple MRI modalities, yet incomplete modality acquisition is common in clinical practice due to protocol heterogeneity and scan failures. Although recent methods maintain segmentation accuracy under missing modality conditions, they frequently overlook prediction reliability, leading to miscalibrated confidence estimates that hinder clinical adoption. Existing calibration techniques are largely modality-agnostic or assume that prediction difficulty decreases monotonically as additional modalities become available. However, in brain tumor segmentation, prediction difficulty depends primarily on which modalities are absent rather than how many, leading to combination-specific and spatially heterogeneous calibration errors. To address this, we propose \textbf{M}issing \textbf{M}odality-\textbf{A}ware \textbf{L}ocal \textbf{T}emperature \textbf{S}caling (\textbf{MMA-LTS}), a post-hoc voxel-wise confidence calibration method. It estimates a spatially adaptive temperature field conditioned on a modality-availability learnable token and a voxel-wise difficulty score. Experiments on BraTS 2020 and FeTS 2024 show that MMA-LTS improves calibration while preserving the segmentation accuracy of state-of-the-art models across diverse missing-modality scenarios, thereby enhancing trustworthiness toward clinical deployment.
\end{abstract}


\section{Introduction}
Multimodal brain tumor segmentation plays a critical role in diagnosis, treatment planning, and prognosis assessment. Accurate delineation requires integrating complementary information from multiple MRI modalities, such as T1, T2, T1ce, and FLAIR~\cite{ding2021rfnet}. However, in real-world clinical workflows, acquiring all modalities is often infeasible due to variations in imaging protocols, scan failures, or patient conditions~\cite{zhang2024tc}. As a result, models must operate robustly under incomplete modality settings. Prior approaches addressed this challenge by synthesizing missing modalities~\cite{meng2024multi,shen2020multi} or training separate models for each possible modality combination~\cite{hu2020knowledge,qiu2023scratch}. More recent methods achieve stronger performance by embedding multimodal inputs into a shared latent space~\cite{chen2019robust,li2025dc,zhang2022mmformer}, enabling a unified architecture.

\begin{figure}[t]
    \centering
    \includegraphics[width=0.98\linewidth]{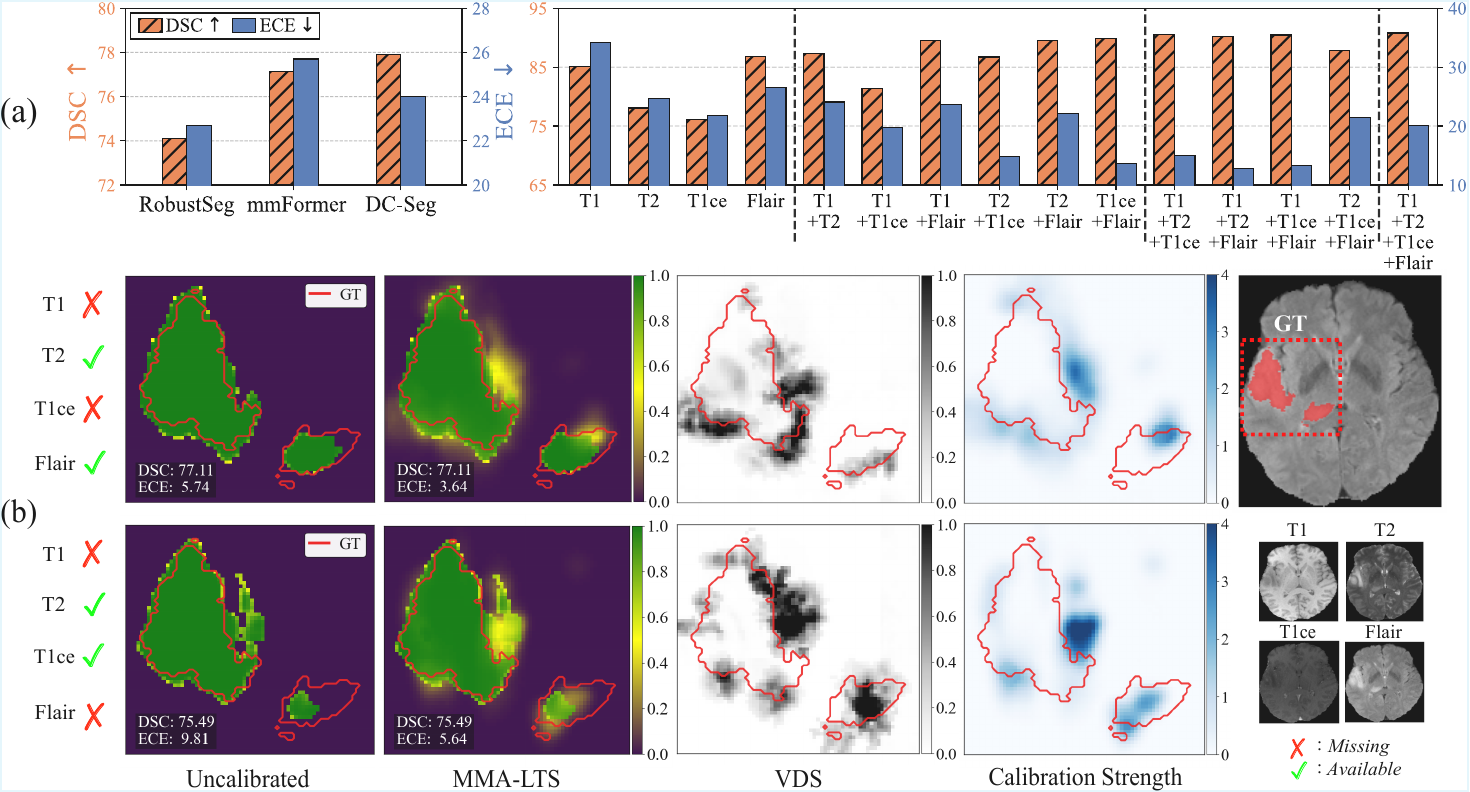}
    \caption{{Segmentation performance and calibration results.} 
    (a) Left: DSC and ECE of state-of-the-art models on BraTS 2020, averaged over all modality combinations. Right: DC-Seg results per modality combination for WT.
    (b) Two cases with the same modality count but differing in combination, showing DC-Seg confidence maps (uncalibrated and MMA-LTS), voxel-wise difficulty score (VDS), and calibration strength ($|T-1|$).}
    \label{fig:figure1}
\end{figure}

Despite recent advances, most studies prioritize segmentation accuracy while paying limited attention to prediction reliability under missing-modality conditions. Fig.~\ref{fig:figure1}(a) shows that the state-of-the-art DC-Seg~\cite{li2025dc} achieves higher Dice similarity coefficient (DSC) yet exhibits larger expected calibration error (ECE) than RobustSeg~\cite{chen2019robust}. This suggests that high segmentation performance under incomplete multimodal inputs can obscure poorly calibrated confidence estimates that do not reflect true accuracy. Reliable confidence is essential in medical image segmentation, as miscalibration can undermine clinical decision-making~\cite{barfoot2024average}. This issue is particularly pronounced in brain tumor segmentation, where heterogeneous subregions and boundaries require calibrated confidence to identify unreliable delineations and support expert intervention~\cite{ding2021local,mehrtash2020confidence}.

Most prior work on segmentation calibration relies on joint prediction-calib
ration methods~\cite{barfoot2024average,liang2025calibrating,mukhoti2020calibrating} or post-processing probability correction~\cite{ding2021local,wang2023calibrating}, which assume fixed and fully observed inputs and therefore do not address multimodal or missing-modality settings. Although calibration methods for multimodal classification under missing modalities have been proposed~\cite{li2025simmlm,ma2023calibrating}, they typically assume prediction difficulty decreases and confidence increases monotonically as more modalities become available. This assumption does not hold in brain tumor segmentation, where modalities contribute unequally: T1ce primarily highlights enhancing tumor (ET) regions, whereas T2 and FLAIR better capture whole tumor (WT) extent~\cite{menze2014multimodal}. As illustrated in Fig.~\ref{fig:figure1}(b), prediction difficulty depends on which modalities are missing rather than how many, resulting in spatially heterogeneous calibration errors. 

\begin{figure}[t]
    \centering
    \includegraphics[width=0.9\linewidth]{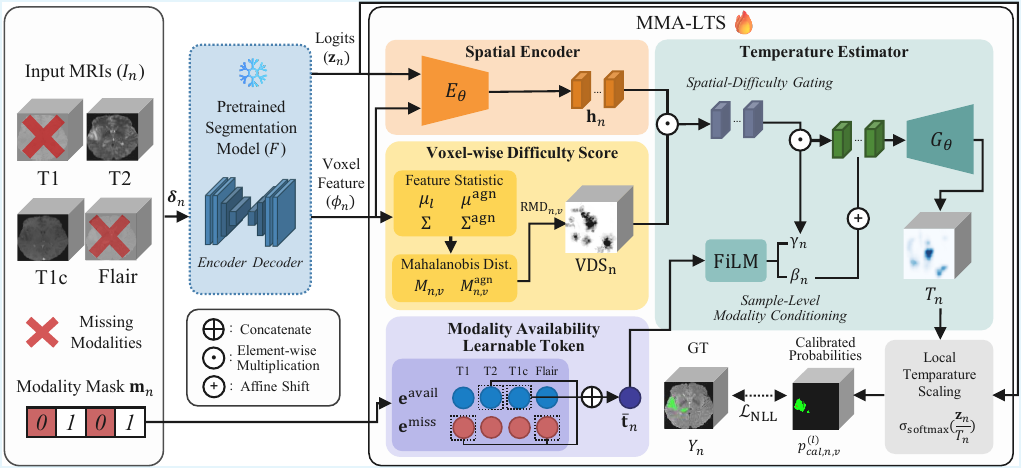}
    \caption{The overall framework of MMA-LTS.}
    \label{fig:figure2}
\end{figure}

Motivated by this observation, we propose \textbf{MMA-LTS} (\textbf{M}issing \textbf{M}odality-\textbf{A}ware \textbf{L}ocal \textbf{T}emperature \textbf{S}caling), a plug-in post-hoc calibration method for multimodal segmentation with incomplete modalities. Unlike prior work, MMA-LTS performs voxel-wise calibration via modality-aware conditioning and spatial difficulty gating, correcting modality-combination-specific and spatially heterogeneous calibration errors without retraining. Experiments on BraTS 2020~\cite{menze2014multimodal} and FeTS 2024~\cite{zhang2024tc} demonstrate improved calibration across missing-modality scenarios while preserving segmentation accuracy.

\section{Method}
Assume disjoint data splits $\mathcal{D}_{\text{train}}$, $\mathcal{D}_{\text{val}}$, and $\mathcal{D}_{\text{test}}$, each consisting of pairs $(I_n, Y_n)$, where $n$ is the sample index, $I_n \in \mathbb{R}^{|\mathcal{M}|\times D\times H\times W}$ is the multimodal input, $Y_n \in L^{D\times H\times W}$ is the ground-truth label, and $\mathcal{M}$ and $L$ denote the sets of modalities and classes, respectively. A segmentation model $F$ pretrained on $\mathcal{D}_{\text{train}}$ takes $I_n$ and produces logits $\mathbf{z}_n \in \mathbb{R}^{|L|\times D\times H\times W}$. The proposed method aims to calibrate $\mathbf{z}_n$ of $F$ on $\mathcal{D}_{\text{test}}$ under arbitrary modality availability. To this end, a calibration module is trained on $\mathcal{D}_{\text{val}}$, yielding spatially adaptive confidence estimates without retraining $F$. The overall framework is shown in Fig.~\ref{fig:figure2}.
\subsection{Missing Modality-Aware Local Temperature Scaling}
\textbf{Spatial Encoder.} $F$ extracts modality-specific features $\{\mathbf{f}_{n,k}\}_{k\in\mathcal{M}}$ and produces a joint representation $\boldsymbol{\phi}_n \in \mathbb{R}^{C_f\times D\times H\times W}$ and $\mathbf{z}_n$ from $I_n$, where $C_f$ is the channel dimension. These two signals are concatenated and passed through a 3D CNN module $E_\theta$ to obtain a spatial embedding $\mathbf{h}_n \in \mathbb{R}^{C_f\times D\times H\times W}$, which serves as the base representation for temperature estimation.
\\
\textbf{Voxel-wise Difficulty Score (VDS).} To capture spatial difficulty induced by modality loss, we extend Mahalanobis-distance-based estimation~\cite{cui2023learning} to 3D segmentation. We collect voxel feature-label pairs $\mathcal{V}=\{(\mathbf{g}_{n,v}, y_{n,v})\}$ from $\mathcal{D}_{\text{train}}$ to model the feature distribution of pretrained $F$, where $v$ indexes a voxel location, $\mathbf{g}_{n,v} = \boldsymbol{\phi}_n(v) \in \mathbb{R}^{C_f}$ denotes the feature vector at voxel $v$, $y_{n,v} = Y_n(v) \in L$ the corresponding label, and $\mathcal{V}_l$ the subset of voxels belonging to each class $l \in L$. We estimate the class-conditional and class-agnostic statistics as:

\begin{equation}
\begin{aligned}
\boldsymbol{\mu}_l &= \frac{1}{|\mathcal{V}_l|} \sum_{(n,v) \in \mathcal{V}_l} \mathbf{g}_{n,v}, \qquad
\boldsymbol{\Sigma} = \frac{1}{|\mathcal{V}|} \sum_{l \in L} \sum_{(n,v)\in \mathcal{V}_l} (\mathbf{g}_{n,v} - \boldsymbol{\mu}_l)(\mathbf{g}_{n,v} - \boldsymbol{\mu}_l)^\top, \\
\boldsymbol{\mu}^{\mathrm{agn}} &= \frac{1}{|\mathcal{V}|} \sum_{(n,v) \in \mathcal{V}} \mathbf{g}_{n,v}, \qquad
\boldsymbol{\Sigma}^{\mathrm{agn}} = \frac{1}{|\mathcal{V}|} \sum_{(n,v)\in \mathcal{V}} (\mathbf{g}_{n,v} - \boldsymbol{\mu}^{\mathrm{agn}}) (\mathbf{g}_{n,v} - \boldsymbol{\mu}^{\mathrm{agn}})^\top.
\end{aligned}
\label{eq:eq5}
\end{equation}
The class-conditional and agnostic Mahalanobis distances are then defined as:
\begin{equation}
\begin{aligned}
M_{n,v}=(\mathbf{g}_{n,v} - \bm{\mu}_{y_{n,v}})^\top 
\bm{\Sigma}^{-1} 
(\mathbf{g}_{n,v} - \bm{\mu}_{y_{n,v}}),  \\
M^{\mathrm{agn}}_{n,v}=(\mathbf{g}_{n,v} - \bm{\mu}_{\mathrm{agn}})^\top 
(\bm{\Sigma}^{\mathrm{agn}})^{-1}
(\mathbf{g}_{n,v} - \bm{\mu}^{\mathrm{agn}}).
\label{eq:eq6}
\end{aligned}
\end{equation}
$M_{n,v}$ measures deviation from the class distribution and $M^{\mathrm{agn}}_{n,v}$ from the global distribution, where $y_{n,v}$ is obtained from ground-truth labels during training and replaced by the predicted label $\hat{y}_{n,v} = \argmax \mathbf{z}_{n,v}$ at inference time. We adopt $\mathrm{RMD}_{n,v} = M_{n,v} - M^{\mathrm{agn}}_{n,v}$ as a voxel-wise difficulty proxy, where larger values indicate voxels that are atypical within their class yet unremarkable globally, reflecting low class discriminability and thus indicating harder regions. After normalization and reshaping, we obtain $\mathrm{VDS}_n \in [0,1]^{D \times H \times W}$.
\\
\textbf{Modality Availability Learnable Token.}
We encode modality availability as a sample-level conditioning signal for calibration.
Let $\mathcal{S}(I_n)\subseteq\mathcal{M}$ denote the set of available modalities and define the binary mask $\mathbf{m}_{n,k}=\mathbb{I}[k\in\mathcal{S}(I_n)]$, where 1 and 0 indicate available and missing, respectively.
For each $k\in\mathcal{M}$, learnable tokens $\mathbf{e}_k^{\text{avail}}, \mathbf{e}_k^{\text{miss}}\in\mathbb{R}^{d_e}$ represent the available and missing states, where $d_e$ is the token dimension.
The modality state token $\mathbf{t}_{n,k}$ is defined as
\begin{equation}
\mathbf{t}_{n,k} = \mathbf{m}_{n,k}\mathbf{e}_k^{\text{avail}} + (1-\mathbf{m}_{n,k})\mathbf{e}_k^{\text{miss}}.
\label{eq:eq3}
\end{equation}
Concatenating all modality tokens yields the summary vector $\bar{\mathbf{t}}_n = \bigoplus_{k \in \mathcal{M}} \mathbf{t}_{n,k} \in \mathbb{R}^{|\mathcal{M}| \cdot d_e}$, which serves to globally modulate the calibration strength.
\\
\textbf{Temperature Estimator.}
The estimator learns a mapping that produces a voxel-wise temperature field using spatial features $\mathbf{h}_{n}$, voxel-wise difficulty score $\mathrm{VDS}_{n}$, and modality availability token $\bar{\mathbf{t}}_n$. Spatial difficulty gating is first applied to selectively emphasize regions requiring stronger calibration. Sample-level modality conditioning is then performed via FiLM-based affine modulation~\cite{perez2018film} to adjust feature modulation according to the sample-wise modality configuration. The resulting temperature field  $T_n \in \mathbb{R}^{D\times H\times W}$ is then given by
\begin{equation}
T_{n} =
G_\theta
\Big(
(\mathbf{h}_{n}\odot \mathrm{VDS}_{n})
\odot \gamma_n
+
\beta_n
\Big),
\end{equation}
where $\odot$ denotes element-wise multiplication, 
$(\gamma_n, \beta_n)=f_{\mathrm{FiLM}}(\bar{\mathbf{t}}_n)$ denote affine modulation parameters,
and $G_\theta$ is a 3D CNN predictor. 
Spatial-difficulty gating determines \textit{{where}} stronger calibration is needed, while sample-level modality conditioning controls \textit{{how strongly}} it is applied based on available modalities.
\\
\textbf{Local Temperature Scaling.} Following prior work~\cite{ding2021local}, the estimated temperature field $T_n$ is applied to post-hoc rescale the logits $\mathbf{z}_{n}$ of $F$ via temperature-scaled softmax, yielding calibrated voxel-wise class probabilities:
\begin{equation}
p_{\mathrm{cal},n,v}^{(l)}
=
\sigma_{\mathrm{softmax}}
\left(
\frac{\mathbf{z}_{n,v}}{T_{n,v}}
\right)^{(l)},\;\;l \in L.
\end{equation}
where $T_{n,v}$ and $\mathbf{z}_{n,v}$ denote the temperature and logit at voxel $v$, respectively. Since $T_{n,v}\in\mathbb{R}^+$, class ordering is preserved, with values greater than 1 reducing overconfidence and values less than 1 mitigating underconfidence.
\\
\textbf{Training Objective.} ModDrop~\cite{neverova2015moddrop} is applied as $\tilde{\mathbf{f}}_{n,k}=\delta_{n,k}\mathbf{f}_{n,k}$, where $\delta_{n,k}$ is a Bernoulli indicator randomly set to 0 during training. MMA-LTS is optimized via the negative log-likelihood (NLL) loss over the voxel grid $\Omega$, which counteracts the miscalibration of $F$ by recovering the true probability distribution:
\begin{equation}
\mathcal{L}_{\mathrm{NLL}} = -\frac{1}{|\Omega|}\sum_{v \in \Omega} \log p_{\mathrm{cal},n,v}^{(y_{n,v})}
\label{eq:eq11}
\end{equation}

\section{Experiments}
\subsection{Implementation Details}
\textbf{Dataset.} Experiments are conducted on BraTS 2020~\cite{menze2014multimodal} and FeTS 2024~\cite{zhang2024tc}, each providing 3D multi-contrast MRI scans with four modalities (FLAIR, T1ce, T1, T2) and three tumor subregions: whole tumor (WT), tumor core (TC), and enhancing tumor (ET). BraTS 2020 contains 369 cases split as (219/50/100) for ($\mathcal{D}_{\text{train}}$/$\mathcal{D}_{\text{val}}$/$\mathcal{D}_{\text{test}}$)~\cite{li2025dc}, and FeTS 2024 contains 1,251 cases split as (1000/125/
126)~\cite{zhu2025bridging}. All volumes are skull-stripped, co-registered to a common anatomical template, resampled to $1mm^3$, and z-score normalized.
\\
\textbf{Segmentation Backbones and Calibration Methods.}
We evaluate MMA-LTS on state-of-the-art brain tumor segmentation models for missing-modality robustness: RobustSeg~\cite{chen2019robust}, mmFormer~\cite{zhang2022mmformer}, and DC-Seg~\cite{li2025dc}. For calibration, we compare against joint prediction-calibration approaches such as mL1-ACE~\cite{barfoot2024average} and SDC~\cite{liang2025calibrating}, as well as post-hoc probability correction methods, including Temperature Scaling (TS)~\cite{guo2017calibration}, Dirichlet Scaling (DirS)~\cite{kull2019beyond}, Local Temperature Scaling (LTS)~\cite{ding2021local}, Selective Scaling (SelS)~\cite{wang2023calibrating}, and MC Dropout (MC)~\cite{gal2016dropout}. All calibration methods are applied uniformly across backbones.
\\
\textbf{Training Settings.} 
All backbones are trained on $\mathcal{D}_{\text{train}}$ with batch size 2 for 500/250 epochs on BraTS 2020/FeTS 2024, augmented with random $80^3$ crops, $\pm10^\circ$ rotation, intensity shifts, and flipping. Post-hoc calibration methods and MMA-LTS are trained for 40 epochs using AdamW~\cite{loshchilov2017decoupled} (batch size $1$, learning rate $1\times10^{-4}$) on $\mathcal{D}_{\text{val}}$, while mL1-ACE and SDC are trained with backbones. MC uses $p=0.3$ with 5 forward passes. For MMA-LTS, we set $C_f=16$ and $d_e=256$. All experiments are conducted on an NVIDIA$^\circledR$ RTX A6000 GPU.
\\
\textbf{Evaluation Metrics.} We evaluate segmentation with DSC~\cite{zijdenbos1994morphometric} and calibration with ECE~\cite{naeini2015obtaining} following recent works~\cite{barfoot2024average,liang2025calibrating}, where ECE is computed over foreground voxels using 10 bins and averaged across test volumes.

\begin{table*}[t]
\centering
\sisetup{round-mode=places, round-precision=1}  
\tiny  
\caption{DSC and ECE comparison of DC-Seg on BraTS 2020. {$\bullet$} and {$\circ$} denote available and missing modalities. Temperature scaling-based methods (TS, DirS, LTS, SelS, MMA-LTS) preserve the DSC of the uncalibrated model; only ECE is reported.}
\begin{tabularx}{\textwidth}{
  >{\centering\arraybackslash}m{0.055\textwidth} | 
  l | 
  *{4}{>{\centering\arraybackslash}X} | 
  *{6}{>{\centering\arraybackslash}X} | 
  *{4}{>{\centering\arraybackslash}X} | 
  >{\centering\arraybackslash}X | 
  >{\centering\arraybackslash}X
}
\toprule
\multirow{4}{*}{{Type}}
& FLAIR
& \scalebox{1.3}{$\circ$} & \scalebox{1.3}{$\circ$} & \scalebox{1.3}{$\circ$} & \scalebox{1.3}{$\bullet$}
& \scalebox{1.3}{$\circ$} & \scalebox{1.3}{$\circ$} & \scalebox{1.3}{$\bullet$} & \scalebox{1.3}{$\circ$}
& \scalebox{1.3}{$\bullet$} & \scalebox{1.3}{$\bullet$} & \scalebox{1.3}{$\bullet$} & \scalebox{1.3}{$\bullet$}
& \scalebox{1.3}{$\bullet$} & \scalebox{1.3}{$\circ$} & \scalebox{1.3}{$\bullet$}
& \multirow{4}{*}{Avg.} \\
& T1   & \scalebox{1.3}{$\circ$} & \scalebox{1.3}{$\circ$} & \scalebox{1.3}{$\bullet$} & \scalebox{1.3}{$\circ$}
       & \scalebox{1.3}{$\circ$} & \scalebox{1.3}{$\bullet$} & \scalebox{1.3}{$\bullet$} & \scalebox{1.3}{$\bullet$}
       & \scalebox{1.3}{$\circ$} & \scalebox{1.3}{$\circ$} & \scalebox{1.3}{$\bullet$} & \scalebox{1.3}{$\bullet$}
       & \scalebox{1.3}{$\circ$} & \scalebox{1.3}{$\bullet$} & \scalebox{1.3}{$\bullet$} & \\
& T1ce & \scalebox{1.3}{$\circ$} & \scalebox{1.3}{$\bullet$} & \scalebox{1.3}{$\circ$} & \scalebox{1.3}{$\circ$}
       & \scalebox{1.3}{$\bullet$} & \scalebox{1.3}{$\bullet$} & \scalebox{1.3}{$\circ$} & \scalebox{1.3}{$\circ$}
       & \scalebox{1.3}{$\circ$} & \scalebox{1.3}{$\bullet$} & \scalebox{1.3}{$\bullet$} & \scalebox{1.3}{$\circ$}
       & \scalebox{1.3}{$\bullet$} & \scalebox{1.3}{$\bullet$} & \scalebox{1.3}{$\bullet$} & \\
& T2   & \scalebox{1.3}{$\bullet$} & \scalebox{1.3}{$\circ$} & \scalebox{1.3}{$\circ$} & \scalebox{1.3}{$\circ$}
       & \scalebox{1.3}{$\bullet$} & \scalebox{1.3}{$\circ$} & \scalebox{1.3}{$\circ$} & \scalebox{1.3}{$\bullet$}
       & \scalebox{1.3}{$\bullet$} & \scalebox{1.3}{$\circ$} & \scalebox{1.3}{$\circ$} & \scalebox{1.3}{$\bullet$}
       & \scalebox{1.3}{$\bullet$} & \scalebox{1.3}{$\bullet$} & \scalebox{1.3}{$\bullet$} & \\
\midrule
\multirow{13}{*}{WT}
& Uncalibrated & \num{24.72} & \num{21.76} & \num{34.20} & \num{26.62} & \num{14.94}
        & \num{19.67} & \num{23.65} & \num{24.09} & \num{22.14} & \num{13.67}
        & \num{13.28} & \num{21.48} & \num{12.83} & \num{15.02} & \num{13.00}
        & \num{20.07} \\
& $\llcorner$ DSC  & \num{85.09} & \num{78.09} & \num{76.18} & \num{86.79} & \num{87.27}
        & \num{81.41} & \num{89.49} & \num{86.74} & \num{89.52} & \num{89.86}
        & \num{90.60} & \num{90.14} & \num{90.47} & \num{87.81} & \num{90.81}
        & \num{86.68} \\
& mL1-ACE~\cite{barfoot2024average} & \num{27.04} & \num{25.44} & \num{39.41} & \num{26.41} & \num{16.37}
        & \num{23.46} & \num{23.69} & \num{25.90} & \num{22.08} & \num{13.97}
        & \num{13.80} & \num{21.24} & \num{13.00} & \num{16.41} & \num{12.98}
        & \num{21.41} \\
& $\llcorner$ DSC & \num{83.73} & \num{76.85} & \num{72.42} & \num{86.12} & \num{86.57}
        & \num{79.41} & \num{88.20} & \num{85.13} & \num{88.48} & \num{89.22}
        & \num{89.91} & \num{89.47} & \num{89.97} & \num{86.83} & \num{90.45}
        & \num{85.52} \\
& SDC~\cite{liang2025calibrating} & \num{22.35} & \num{21.75} & \num{33.42} & \num{23.46} & \num{13.72}
        & \num{19.18} & \num{20.42} & \num{21.81} & \num{18.99} & \num{12.67}
        & \num{12.22} & \num{18.35} & \num{11.68} & \num{13.83} & \num{11.58}
        & \num{18.36} \\
& $\llcorner$ DSC & \num{85.63} & \num{78.34} & \num{75.56} & \num{86.81} & \num{87.73}
        & \num{81.46} & \num{89.54} & \num{87.05} & \num{89.76} & \num{90.07}
        & \num{90.75} & \num{90.38} & \num{90.71} & \num{88.11} & \num{91.00}
        & \num{86.86} \\
& MC~\cite{gal2016dropout}   & \num{24.78} & \num{21.95} & \num{34.17} & \num{26.72} & \num{14.93}
        & \num{19.74} & \num{23.55} & \num{24.30} & \num{22.19} & \num{13.57}
        & \num{13.39} & \num{21.52} & \num{12.93} & \num{15.25} & \num{12.96}
        & \num{20.13} \\
& $\llcorner$ DSC  & \num{84.96} & \num{77.90} & \num{75.96} & \num{86.74} & \num{87.28}
        & \num{81.22} & \num{89.41} & \num{86.44} & \num{89.30} & \num{89.84}
        & \num{90.60} & \num{90.20} & \num{90.42} & \num{87.65} & \num{90.89}
        & \num{86.59} \\
& TS \cite{guo2017calibration} & \num{23.85} & \num{21.05} & \num{33.16} & \num{25.76} & \num{14.33}
        & \num{18.97} & \num{22.84} & \num{23.24} & \num{21.35} & \num{13.11}
        & \num{12.74} & \num{20.72} & \num{12.28} & \num{14.41} & \num{12.46}
        & \num{19.35} \\
& DirS~\cite{kull2019beyond}  & \num{24.89} & \num{22.90} & \num{34.73} & \num{26.57} & \num{15.62}
        & \num{20.59} & \num{23.38} & \num{24.32} & \num{22.07} & \num{14.11}
        & \num{13.53} & \num{21.35} & \num{13.25} & \num{15.61} & \num{13.22}
        & \num{20.41} \\
& LTS~\cite{ding2021local}   &  {\num{14.23}} &  {\num{16.23}} &  {\num{22.91}} &  {\num{16.24}} &  {\num{10.28}}
        &  {\num{14.24}} &  {\num{14.00}} &  {\num{13.93}} &  {\num{13.07}} &  {\num{10.19}}
        &  {\num{9.73}}  &  {\num{12.35}} &  {\num{9.47}}  &  {\num{10.38}} &  {\num{9.35}}
        &  {\num{13.11}} \\
& SelS~\cite{wang2023calibrating} & \num{20.43} & \num{18.17} & \num{29.33} & \num{22.93} & \num{11.27}
        & \num{15.20} & \num{19.91} & \num{19.83} & \num{18.65} & \num{10.65}
        & \num{10.40} & \num{18.05} & \num{9.94}  & \num{11.23} & \num{10.14}
        & \num{16.29} \\
& \hl{MMA-LTS$^\ast$}
& \hl{\textbf{\num{12.64}}} & \hl{\textbf{\num{12.74}}} & \hl{\textbf{\num{15.08}}} & \hl{\textbf{\num{12.99}}} & \hl{\textbf{\num{8.11}}}
& \hl{\textbf{\num{10.22}}} & \hl{\textbf{\num{11.77}}} & \hl{\textbf{\num{11.18}}} & \hl{\textbf{\num{11.44}}} & \hl{\textbf{\num{8.17}}}
& \hl{\textbf{\num{7.22}}}  & \hl{\textbf{\num{10.12}}} & \hl{\textbf{\num{6.95}}}  & \hl{\textbf{\num{7.3}}}  & \hl{\textbf{\num{6.77}}}
& \hl{\textbf{\num{10.18}}} \\
\rowcolor[HTML]{EFEFEF}
& Uncalibrated
& \num{37.56} & \num{19.06} & \num{43.16} & \num{43.39} & \num{16.53}
& \num{17.91} & \num{38.89} & \num{35.61} & \num{35.49} & \num{16.42}
& \num{17.40} & \num{35.38} & \num{16.08} & \num{16.94} & \num{17.01}
& \num{27.12} \\
\rowcolor[HTML]{EFEFEF}
& $\llcorner$ DSC
& \num{73.24} & \num{84.52} & \num{64.60} & \num{70.67} & \num{87.58}
& \num{86.83} & \num{73.93} & \num{73.85} & \num{75.31} & \num{87.36}
& \num{87.40} & \num{75.51} & \num{87.48} & \num{87.20} & \num{87.34}
& \num{80.19} \\
\rowcolor[HTML]{EFEFEF}
& mL1-ACE & \num{33.60} & \num{17.64} & \num{40.73} & \num{38.83} & \num{13.98}
        & \num{16.63} & \num{34.28} & \num{32.23} & \num{31.05} & \num{13.35}
        & \num{14.22} & \num{30.65} & \num{12.90} & \num{14.31} & \num{13.78}
        & \num{23.88} \\
\rowcolor[HTML]{EFEFEF}
& $\llcorner$ DSC & \num{72.45} & \num{84.01} & \num{63.22} & \num{70.02} & \num{87.35}
        & \num{84.86} & \num{72.81} & \num{73.01} & \num{74.43} & \num{87.54}
        & \num{87.38} & \num{74.89} & \num{87.55} & \num{86.73} & \num{87.13}
        & \num{79.56} \\
\rowcolor[HTML]{EFEFEF} 
& SDC & \num{31.81} & \num{16.73} & \num{37.63} & \num{36.25} & \num{14.03}
        & \num{15.01} & \num{32.20} & \num{30.13} & \num{29.11} & \num{13.99}
        & \num{14.68} & \num{29.08} & \num{13.48} & \num{14.49} & \num{14.36}
        & \num{22.87} \\
\rowcolor[HTML]{EFEFEF}        
& $\llcorner$ DSC & \num{73.27} & \num{84.65} & \num{64.30} & \num{70.51} & \num{87.56}
        & \num{86.04} & \num{74.13} & \num{73.99} & \num{75.84} & \num{87.06}
        & \num{87.10} & \num{76.01} & \num{87.24} & \num{86.92} & \num{87.01}
        & \num{80.11} \\
\rowcolor[HTML]{EFEFEF}
TC & MC
& \num{37.56} & \num{19.39} & \num{42.84} & \num{43.65} & \num{16.64}
& \num{17.81} & \num{38.56} & \num{35.79} & \num{35.57} & \num{16.39}
& \num{17.50} & \num{35.29} & \num{15.98} & \num{17.00} & \num{17.04}
& \num{27.13} \\
\rowcolor[HTML]{EFEFEF}
& $\llcorner$ DSC & \num{72.99} & \num{84.83} & \num{54.84} & \num{70.70} & \num{87.56}
        & \num{86.93} & \num{74.33} & \num{73.70} & \num{75.38} & \num{87.47}
        & \num{87.50} & \num{75.45} & \num{87.46} & \num{87.35} & \num{87.37}
        & \num{80.26} \\
\rowcolor[HTML]{EFEFEF}
& TS
& \num{36.47} & \num{18.41} & \num{41.91} & \num{42.22} & \num{15.93}
& \num{17.24} & \num{37.79} & \num{34.56} & \num{34.44} & \num{15.78}
& \num{16.76} & \num{34.36} & \num{15.47} & \num{16.32} & \num{16.39}
& \num{26.27} \\
\rowcolor[HTML]{EFEFEF}
& DirS
& \num{34.86} & \num{17.64} & \num{40.48} & \num{40.67} & \num{15.16}
& \num{16.30} & \num{36.05} & \num{32.97} & \num{32.85} & \num{15.07}
& \num{15.81} & \num{32.73} & \num{14.75} & \num{15.46} & \num{15.52}
& \num{25.09} \\
\rowcolor[HTML]{EFEFEF}
& LTS
& \num{22.16} & \num{12.58} & \num{27.48} & \num{27.72} & \num{10.60}
& \num{11.08} & \num{24.10} & \num{21.18} & \num{21.17} & \num{10.16}
& \num{10.52} & \num{21.59} & \num{10.04} & \num{10.90} & \num{10.67}
& \num{16.80} \\
\rowcolor[HTML]{EFEFEF}
& SelS
& \num{32.16} & \num{15.65} & \num{37.20} & \num{39.78} & \num{14.32}
& \num{15.46} & \num{35.11} & \num{31.98} & \num{32.15} & \num{13.96}
& \num{14.98} & \num{31.95} & \num{13.85} & \num{14.66} & \num{13.48}
& \num{24.16} \\
\rowcolor[HTML]{EFEFEF}
& \hl{MMA-LTS$^\ast$}
& \hl{\textbf{\num{18.15}}} & \hl{\textbf{\num{10.86}}} & \hl{\textbf{\num{20.39}}} & \hl{\textbf{\num{20.07}}} & \hl{\textbf{\num{10.2}}}
& \hl{\textbf{\num{10.05}}} & \hl{\textbf{\num{20.94}}} & \hl{\textbf{\num{18.88}}} & \hl{\textbf{\num{17.58}}} & \hl{\textbf{\num{9.68}}}
& \hl{\textbf{\num{10.33}}} & \hl{\textbf{\num{19.34}}} & \hl{\textbf{\num{9.7}}}  & \hl{\textbf{\num{10.61}}} & \hl{\textbf{\num{10.62}}}
& \hl{\textbf{\num{14.49}}} \\
\multirow{13}{*}{ET}
& Uncalibrated
& \num{36.45} & \num{12.80} & \num{46.84} & \num{46.89} & \num{9.90}
& \num{10.88} & \num{43.36} & \num{37.22} & \num{37.47} & \num{11.16}
& \num{10.47} & \num{37.96} & \num{10.43} & \num{10.06} & \num{10.35}
& \num{24.82} \\
& $\llcorner$ DSC
& \num{51.76} & \num{79.99} & \num{40.92} & \num{42.42} & \num{83.24}
& \num{83.35} & \num{47.02} & \num{51.60} & \num{52.45} & \num{82.67}
& \num{83.51} & \num{52.59} & \num{83.17} & \num{83.59} & \num{83.64}
& \num{66.80} \\
& mL1-ACE & \num{31.83} & \num{11.09} & \num{45.31} & \num{42.98} & \num{7.77}
        & \num{10.45} & \num{39.55} & \num{34.40} & \num{32.88} & \num{8.47}
        & \num{7.80} & \num{33.58} & \num{7.90} & \num{7.96} & \num{7.93}
        & \num{21.99} \\
& $\llcorner$ DSC & \num{51.39} & \num{79.35} & \num{38.56} & \num{41.78} & \num{83.22}
        & \num{81.21} & \num{45.82} & \num{50.25} & \num{51.87} & \num{82.74}
        & \num{83.52} & \num{51.97} & \num{83.12} & \num{83.27} & \num{83.32}
        & \num{66.09} \\
& SDC   & \num{11.80} & \num{31.52} & \num{42.71} & \num{38.02} & \num{8.85} & \num{9.70}
        & \num{36.18} & \num{32.59} & \num{31.39} & \num{9.48} & \num{8.93}
        & \num{32.12} & \num{8.85} & \num{8.88} & \num{9.00}
        & \num{21.33} \\
& $\llcorner$ DSC & \num{51.30} & \num{80.00} & \num{39.07} & \num{43.97} & \num{83.30}
        & \num{83.32} & \num{47.69} & \num{50.96} & \num{54.66} & \num{82.77}
        & \num{83.48} & \num{52.69} & \num{83.23} & \num{83.48} & \num{83.43}
        & \num{83.48} \\
& MC
& \num{36.37} & \num{13.06} & \num{46.58} & \num{47.57} & \num{10.09}
& \num{10.54} & \num{42.99} & \num{37.40} & \num{37.56} & \num{11.14}
& \num{10.61} & \num{37.67} & \num{10.29} & \num{10.03} & \num{10.28}
& \num{24.81} \\
& $\llcorner$ DSC & \num{51.41} & \num{80.75} & \num{40.52} & \num{41.36} & \num{83.39}
        & \num{83.84} & \num{46.68} & \num{51.06} & \num{51.81} & \num{83.16}
        & \num{83.61} & \num{52.28} & \num{83.68} & \num{83.95} & \num{83.74}
        & \num{66.75} \\
& TS
& \num{35.30} & \num{12.33} & \num{45.54} & \num{45.66} & \num{9.46}
& \num{10.38} & \num{42.21} & \num{36.08} & \num{36.38} & \num{10.67}
& \num{9.99} & \num{36.88} & \num{9.96} & \num{9.59} & \num{9.87}
& \num{24.02} \\
& DirS
& \num{33.26} & \num{11.96} & \num{43.47} & \num{43.47} & \num{9.08}
& \num{10.00} & \num{39.69} & \num{33.94} & \num{34.22} & \num{10.28}
& \num{9.43} & \num{34.63} & \num{9.53} & \num{9.09} & \num{9.30}
& \num{22.76} \\
& LTS
& \num{21.45} & \num{9.36} & \num{30.61} & \num{31.70} & \num{7.47}
& \num{7.60} & \num{29.28} & \textbf{\num{22.89}} & \num{23.40} & \num{8.12}
& \num{7.70} & \num{24.42} & \num{7.74} & \num{7.58} & \num{7.71}
& \num{16.47} \\
& SelS
& \num{30.87} & \num{11.21} & \num{40.38} & \num{42.73} & \num{8.59}
& \num{9.12} & \num{39.15} & \num{32.92} & \num{33.51} & \num{9.76}
& \num{9.20} & \num{33.92} & \num{9.27} & \num{8.72} & \num{9.09}
& \num{22.10} \\
& \hl{MMA-LTS$^\ast$}
& \hl{\textbf{\num{21.14}}} & \hl{\textbf{\num{8.84}}} & \hl{\textbf{\num{25.1}}} & \hl{\textbf{\num{26.99}}} & \hl{\textbf{\num{7.14}}}
& \hl{\textbf{\num{7.00}}} & \hl{\textbf{\num{27.97}}} & \hl{\textbf{\num{22.88}}} & \hl{\textbf{\num{21.93}}} & \hl{\textbf{\num{7.78}}}
& \hl{\textbf{\num{6.85}}} & \hl{\textbf{\num{23.36}}} & \hl{\textbf{\num{6.82}}} & \hl{\textbf{\num{6.62}}} & \hl{\textbf{\num{6.44}}}
& \hl{\textbf{\num{15.12}}} \\
\bottomrule
\end{tabularx}
\label{tab:tab1}
\end{table*}

\begin{table*}[t]
\centering
\tiny
\sisetup{round-mode=places, round-precision=1}
\caption{{DSC and ECE comparison of DC-Seg on FeTS 2024.}}
\begin{tabularx}{\textwidth}{
  >{\centering\arraybackslash}m{0.055\textwidth} | 
  l | 
  *{4}{>{\centering\arraybackslash}X} | 
  *{6}{>{\centering\arraybackslash}X} | 
  *{4}{>{\centering\arraybackslash}X} | 
  >{\centering\arraybackslash}X | 
  >{\centering\arraybackslash}X
}
\toprule
\multirow{4}{*}{{Type}}
& FLAIR
& \scalebox{1.3}{$\circ$} & \scalebox{1.3}{$\circ$} & \scalebox{1.3}{$\circ$} & \scalebox{1.3}{$\bullet$}
& \scalebox{1.3}{$\circ$} & \scalebox{1.3}{$\circ$} & \scalebox{1.3}{$\bullet$} & \scalebox{1.3}{$\circ$}
& \scalebox{1.3}{$\bullet$} & \scalebox{1.3}{$\bullet$} & \scalebox{1.3}{$\bullet$} & \scalebox{1.3}{$\bullet$}
& \scalebox{1.3}{$\bullet$} & \scalebox{1.3}{$\circ$} & \scalebox{1.3}{$\bullet$}
& \multirow{4}{*}{Avg.} \\
& T1   & \scalebox{1.3}{$\circ$} & \scalebox{1.3}{$\circ$} & \scalebox{1.3}{$\bullet$} & \scalebox{1.3}{$\circ$}
       & \scalebox{1.3}{$\circ$} & \scalebox{1.3}{$\bullet$} & \scalebox{1.3}{$\bullet$} & \scalebox{1.3}{$\bullet$}
       & \scalebox{1.3}{$\circ$} & \scalebox{1.3}{$\circ$} & \scalebox{1.3}{$\bullet$} & \scalebox{1.3}{$\bullet$}
       & \scalebox{1.3}{$\circ$} & \scalebox{1.3}{$\bullet$} & \scalebox{1.3}{$\bullet$} & \\
& T1ce & \scalebox{1.3}{$\circ$} & \scalebox{1.3}{$\bullet$} & \scalebox{1.3}{$\circ$} & \scalebox{1.3}{$\circ$}
       & \scalebox{1.3}{$\bullet$} & \scalebox{1.3}{$\bullet$} & \scalebox{1.3}{$\circ$} & \scalebox{1.3}{$\circ$}
       & \scalebox{1.3}{$\circ$} & \scalebox{1.3}{$\bullet$} & \scalebox{1.3}{$\bullet$} & \scalebox{1.3}{$\circ$}
       & \scalebox{1.3}{$\bullet$} & \scalebox{1.3}{$\bullet$} & \scalebox{1.3}{$\bullet$} & \\
& T2   & \scalebox{1.3}{$\bullet$} & \scalebox{1.3}{$\circ$} & \scalebox{1.3}{$\circ$} & \scalebox{1.3}{$\circ$}
       & \scalebox{1.3}{$\bullet$} & \scalebox{1.3}{$\circ$} & \scalebox{1.3}{$\circ$} & \scalebox{1.3}{$\bullet$}
       & \scalebox{1.3}{$\bullet$} & \scalebox{1.3}{$\circ$} & \scalebox{1.3}{$\circ$} & \scalebox{1.3}{$\bullet$}
       & \scalebox{1.3}{$\bullet$} & \scalebox{1.3}{$\bullet$} & \scalebox{1.3}{$\bullet$} & \\
\midrule
\multirow{13}{*}{WT}
& Uncalibrated
& \num{20.89} & \num{17.00} & \num{25.25} & \num{17.92}
& \num{12.09} & \num{14.66} & \num{15.71} & \num{19.30}
& \num{15.33} & \num{8.69}  & \num{8.49}  & \num{14.73}
& \num{8.49}  & \num{12.00} & \num{14.60}
& \num{15.01} \\
& $\llcorner$ DSC 
& \num{87.65} & \num{79.33} & \num{80.80} & \num{90.31}
& \num{88.97} & \num{83.44} & \num{91.84} & \num{88.53}
& \num{91.94} & \num{91.64} & \num{92.27} & \num{92.22}
& \num{92.50} & \num{89.20} & \num{88.88}
& \num{88.63} \\
& mL1-ACE \cite{barfoot2024average} & \num{20.44} & \num{20.70} & \num{27.20} & \num{15.92} & \num{12.70} 
        & \num{17.43} & \num{14.00} & \num{18.53} & \num{13.84} & \num{8.12} 
        & \num{8.12} & \num{13.18} & \num{7.88} & \num{12.46} & \num{7.96}
        & \num{14.57}
        \\
& $\llcorner$ DSC & \num{86.46} & \num{78.82} & \num{78.44} & \num{90.46} & \num{88.08}
        & \num{82.34} & \num{91.74} & \num{87.34} & \num{91.87} & \num{92.08}
        & \num{92.43} & \num{92.05} & \num{92.61} & \num{88.35} & \num{92.61}
        & \num{88.38} \\
& SDC & \num{18.19} & \num{17.26} & \num{22.60} & \num{15.23} & \num{10.86}
        & \num{13.61} & \num{13.14} & \num{16.30} & \num{13.15} & \num{7.59}
        & \num{7.48} & \num{12.25} & \num{7.34} & \num{10.74} & \num{7.31}
        & \num{10.74} \\
& $\llcorner$ DSC & \num{87.84} & \num{80.11} & \num{80.50} & \num{90.78} & \num{89.16}
        & \num{83.67} & \num{92.09} & \num{88.69} & \num{92.25} & \num{92.10}
        & \num{92.57} & \num{92.41} & \num{92.84} & \num{89.30} & \num{92.78}
        & \num{89.14} \\
& MC \cite{gal2016dropout}
& \num{20.76} & \num{16.91} & \num{25.32} & \num{17.93}
& \num{12.03} & \num{14.69} & \num{15.76} & \num{19.26}
& \num{15.33} & \num{8.68}  & \num{8.52}  & \num{14.71}
& \num{8.50}  & \num{11.98} & \num{14.59}
& \num{15.00} \\
& $\llcorner$ DSC & \num{87.68} & \num{79.28} & \num{80.63} & \num{90.34}
        & \num{88.99} & \num{83.39} & \num{91.97} & \num{88.43} & \num{92.07}
        & \num{91.79} & \num{92.37} & \num{92.37} & \num{92.56} & \num{89.08} & \num{92.70} 
        & \num{88.91} \\
& TS \cite{guo2017calibration}
& \num{18.43} & \num{15.33} & \num{22.21} & \num{15.58}
& \num{10.77} & \num{13.11} & \num{13.48} & \num{16.90}
& \num{13.22} & \num{7.59}  & \num{7.40}  & \num{12.65}
& \num{7.39}  & \num{10.73} & \num{12.81}
& \num{13.17} \\
& DirS \cite{kull2019beyond}
& \num{19.85} & \num{18.02} & \num{25.48} & \num{16.25}
& \num{12.70} & \num{15.55} & \num{14.28} & \num{18.47}
& \num{13.99} & \num{8.49}  & \num{8.47}  & \num{13.50}
& \num{8.48}  & \num{12.60} & \num{14.31}
& \num{14.70} \\
& LTS \cite{ding2021local}
& \num{12.19} & \num{14.01} & \num{15.89} & \num{12.09}
& \num{9.81}  & \num{11.68} & \num{10.04} & \num{11.77}
& \num{9.90}  & \num{8.78}  & \num{8.64}  & \num{9.69}
& \num{8.75}  & \num{10.04} & \num{8.74}
& \num{10.86} \\
& SelS \cite{wang2023calibrating}
& \num{15.07} & \num{13.57} & \num{17.79} & \num{13.02}
& \num{9.14}  & \num{11.95} & \num{11.18} & \num{13.68}
& \num{11.05} & \num{6.63}  & {\num{6.91}}  & \num{10.68}
& \num{7.02}  & \num{8.98}  & \num{10.85}
& \num{11.17} 
\\
& \hl{MMA-LTS$^\ast$}
& \hl{\textbf{\num{11.78}}} & \hl{\textbf{\num{13.23}}} & \hl{\textbf{\num{15.65}}} & \hl{\textbf{\num{11.87}}}
& \hl{\textbf{\num{8.51}}}  & \hl{\textbf{\num{10.30}}} & \hl{\textbf{\num{9.89}}}  & \hl{\textbf{\num{10.26}}}
& \hl{\textbf{\num{8.72}}}  & \hl{\textbf{\num{6.90}}}  & \hl{\textbf{\num{6.49}}}  & \hl{\textbf{\num{8.06}}}
& \hl{\textbf{\num{6.28}}}  & \hl{\textbf{\num{8.44}}}  & \hl{\textbf{\num{6.11}}}  & \hl{\textbf{\num{9.50}}} \\
\rowcolor[HTML]{EFEFEF}
& Uncalibrated
& \num{30.23} & \num{14.46} & \num{32.40} & \num{31.80}
& \num{10.90} & \num{12.96} & \num{26.26} & \num{25.94}
& \num{25.35} & \num{11.61} & \num{11.48} & \num{23.67}
& \num{10.59} & \num{10.83} & \num{19.27}
& \num{19.85} \\
\rowcolor[HTML]{EFEFEF}
& $\llcorner$ DSC 
& \num{75.75} & \num{88.89} & \num{73.83} & \num{76.86}
& \num{91.56} & \num{90.50} & \num{80.39} & \num{80.36}
& \num{81.02} & \num{91.33} & \num{91.81} & \num{82.30}
& \num{92.01} & \num{92.15} & \num{85.40}
& \num{84.29} \\
\rowcolor[HTML]{EFEFEF}
& mL1-ACE & \num{26.49} & \num{13.94} & \num{29.27} & \num{27.18} & \num{9.87}
        & \num{12.11} & \num{21.98} & \num{22.90} & \num{23.19} & \num{10.66}
        & \num{10.31} & \num{21.34} & \num{9.92} & \num{10.36} & \num{9.70}
        & \num{17.28} \\
\rowcolor[HTML]{EFEFEF}
& $\llcorner$ DSC  & \num{74.28} & \num{88.04} & \num{71.43} & \num{75.53} & \num{90.85}
        & \num{89.36} & \num{78.73} & \num{77.70} & \num{78.67} & \num{90.56}
        & \num{90.96} & \num{79.96} & \num{91.12}
        & \num{90.56}  & \num{91.25}  & \num{84.93} \\
\rowcolor[HTML]{EFEFEF}
& SDC & \num{23.42} & \num{12.49} & \num{28.67} & \num{27.63} & \num{9.66}
        & \num{11.10} & \num{22.87} & \num{21.94} & \num{21.90} & \num{10.03}
        & \num{10.21} & \num{20.29} & \num{9.53} & \num{9.69} & \num{9.52}
        & \num{16.73} \\
\rowcolor[HTML]{EFEFEF}
& $\llcorner$ DSC & \num{76.18} & \num{89.19} & \num{73.13} & \num{77.08} & \num{91.60}
        & \num{90.58} & \num{80.19} & \num{80.09} & \num{80.93} & \num{91.67}
        & \num{91.90} & \num{82.30} & \num{91.91} & \num{92.07} & \num{92.22}
        & \num{85.40} \\
\rowcolor[HTML]{EFEFEF}
TC & MC
& \num{30.12} & \num{14.35} & \num{32.72} & \num{31.66}
& \num{10.81} & \num{12.87} & \num{26.28} & \num{25.70}
& \num{25.15} & \num{11.66} & \num{11.45} & \num{23.57}
& \num{10.61} & \num{10.86} & \num{19.22}
& \num{19.80} \\
\rowcolor[HTML]{EFEFEF}
& $\llcorner$ DSC & \num{75.89} & \num{88.91} & \num{73.79} & \num{76.90} & \num{91.500}
        & \num{90.41} & \num{80.29} & \num{80.45} & \num{81.22} & \num{91.36}
        & \num{91.85} & \num{82.68} & \num{92.00} & \num{92.13} & \num{92.38}
        & \num{85.45} \\
\rowcolor[HTML]{EFEFEF}
& TS
& \num{26.86} & \num{13.31} & \num{28.59} & \num{28.41}
& \num{9.80}  & \num{11.85} & \num{23.13} & \num{22.70}
& \num{22.38} & \num{10.50} & \num{10.38} & \num{20.83}
& \num{9.49}  & \num{9.73}  & \num{17.16}
& \num{17.67} \\
\rowcolor[HTML]{EFEFEF} 
& DirS
& \num{26.67} & \num{13.23} & \num{28.48} & \num{28.20}
& \num{9.85}  & \num{11.65} & \num{23.00} & \num{22.68}
& \num{22.14} & \num{10.34} & \num{10.25} & \num{20.69}
& \num{9.39}  & \num{9.59}  & \num{17.04}
& \num{17.55} \\
\rowcolor[HTML]{EFEFEF}
& LTS
& \num{18.18} & \num{14.38} & \num{19.88} & \num{17.69}
& \num{11.66} & \num{13.18} & \num{15.36} & \num{14.50}
& \num{14.74} & \num{11.87} & \num{11.71} & \num{14.16}
& \num{11.28} & \num{11.42} & \num{11.39}
& \num{14.09} \\
\rowcolor[HTML]{EFEFEF}
& SelS
& \num{23.68} & \num{12.47} & \num{23.28} & \num{24.27}
& \num{9.22}  & \num{11.56} & \num{20.27} & \num{19.52}
& \num{20.67} & \num{9.83}  & \num{9.79}  & \num{18.92}
& \num{9.18}  & \num{9.17}  & \num{15.76}
& \num{15.84} \\
\rowcolor[HTML]{EFEFEF}
& \hl{MMA-LTS$^\ast$}
& \hl{\textbf{\num{17.73}}} & \hl{\textbf{\num{11.30}}} & \hl{\textbf{\num{19.50}}} & \hl{\textbf{\num{17.56}}}
& \hl{\textbf{\num{7.83}}}  & \hl{\textbf{\num{9.35}}}  & \hl{\textbf{\num{15.18}}} & \hl{\textbf{\num{14.05}}}
& \hl{\textbf{\num{14.26}}} & \hl{\textbf{\num{8.14}}}  & \hl{\textbf{\num{8.03}}} & \hl{\textbf{\num{13.56}}}
& \hl{\textbf{\num{7.53}}}  & \hl{\textbf{\num{7.55}}}  & \hl{\textbf{\num{7.78}}} & \hl{\textbf{\num{11.96}}} \\
\multirow{13}{*}{ET}
& Uncalibrated
& \num{27.35} & \num{10.15} & \num{31.38} & \num{27.81}
& \num{7.84}  & \num{8.92}  & \num{24.20} & \num{23.61}
& \num{22.56} & \num{7.67}  & \num{7.51}  & \num{21.47}
& \num{7.57}  & \num{7.81}  & \num{16.22}
& \num{16.80} \\
& $\llcorner$ DSC 
& \num{58.88} & \num{85.51} & \num{54.64} & \num{59.57}
& \num{87.63} & \num{87.55} & \num{63.54} & \num{63.67}
& \num{65.25} & \num{87.65} & \num{88.64} & \num{66.82}
& \num{88.02} & \num{88.29} & \num{75.61}
& \num{74.48} \\
& mL1-ACE & \num{23.84} & \num{10.66} & \num{28.74} & \num{24.02} & \num{7.51} 
        & \num{8.90} & \num{20.37} & \num{20.96} & \num{21.05} & \num{7.43} 
        & \num{6.90} & \num{19.52} & \num{6.92} & \num{7.68} & \num{6.53}
        & \num{14.73} \\
& $\llcorner$ DSC & \num{57.95} & \num{84.56} & \num{52.56} & \num{58.14} & \num{87.08}
        & \num{86.34} & \num{62.30} & \num{61.37} & \num{63.82} & \num{87.39}
        & \num{88.14} & \num{65.51} & \num{87.80}
        & \num{86.93}  & \num{88.24}  & \num{74.54} \\
& SDC & \num{23.09} & \num{9.01} & \num{27.76} & \num{23.80} & \num{6.73}
        & \num{7.73} & \num{20.95} & \num{19.78} & \num{19.52} & \num{6.87}
        & \num{6.80} & \num{18.47} & \num{6.74} & \num{6.71} & \num{6.72}
        & \num{14.05} \\
& $\llcorner$ DSC & \num{59.09} & \num{85.46} & \num{54.13} & \num{59.28} & \num{87.74}
        & \num{87.50} & \num{63.44} & \num{63.42} & \num{65.20} & \num{87.79}
        & \num{88.60} & \num{66.85} & \num{88.09} & \num{88.36} & \num{88.46}
        & \num{75.56} \\
& MC
& \num{27.51} & \num{10.12} & \num{31.69} & \num{27.72}
& \num{7.74}  & \num{8.80}  & \num{24.30} & \num{23.48}
& \num{22.49} & \num{7.73}  & \num{7.48}  & \num{21.29}
& \num{7.64}  & \num{7.81}  & \num{16.22}
& \num{16.80} \\
& $\llcorner$ DSC & \num{58.86} & \num{85.53} & \num{54.00} & \num{59.26} & \num{87.53}
        & \num{87.50} & \num{63.04} & \num{63.59} & \num{65.26} & \num{87.71}
        & \num{88.61} & \num{66.89} & \num{88.03}
        & \num{88.20}  & \num{88.54}  & \num{75.50} \\
& TS
& \num{24.17} & \num{9.15}  & \num{27.62} & \num{24.62}
& \num{6.89}  & \num{7.97}  & \num{21.15} & \num{20.54}
& \num{19.75} & \num{6.75}  & \num{6.57}  & \num{18.80}
& \num{6.63}  & \num{6.82}  & \num{14.26}
& \num{14.78} \\
& DirS
& \num{24.33} & \num{9.32}  & \num{27.81} & \num{24.67}
& \num{7.14}  & \num{7.98}  & \num{21.36} & \num{20.90}
& \num{19.79} & \num{6.77}  & \num{6.61}  & \num{18.92}
& \num{6.70}  & \num{6.89}  & \num{14.38}
& \num{14.90} \\
& LTS
& \num{20.29} & \num{11.35} & {\num{21.75}} & \textbf{{\num{18.56}}}
& \num{9.86}  & \num{9.63}  & {\num{17.86}} & \textbf{\num{16.34}}
& \num{16.49} & \num{8.78}  & \num{8.68}  & \num{17.03}
& \num{8.58}  & \num{8.56}  & \num{8.34}
& \num{13.47} \\
& SelS
& \num{24.37} & \num{8.78}  & \num{24.83} & \num{23.96}
& \num{6.45}  & \num{7.68}  & \num{20.04} & \num{20.23}
& \num{18.98} & \num{6.49}  & \num{6.38}  & \num{18.71}
& \num{6.18}  & \num{6.46}  & \num{13.83}
& \num{13.05} \\
& \hl{MMA-LTS$^\ast$}
& \hl{\textbf{\num{20.10}}} & \hl{\textbf{\num{8.62}}} & \hl{\textbf{\num{21.65}}} & \hl{\underline{\num{19.14}}}
& \hl{\textbf{\num{5.92}}} & \hl{\textbf{\num{6.68}}} & \hl{\textbf{\num{17.83}}} & \hl{\underline{\num{16.84}}}
& \hl{\textbf{\num{15.62}}} & \hl{\textbf{\num{5.25}}} & \hl{\textbf{\num{5.06}}} & \hl{\textbf{\num{15.26}}}
& \hl{\textbf{\num{5.03}}} & \hl{\textbf{\num{5.50}}} & \hl{\textbf{\num{5.08}}} & \hl{\textbf{\num{11.57}}} \\
\bottomrule
\end{tabularx}
\label{tab:tab2}
\end{table*}

\begin{table}[t!]
\sisetup{round-mode=places, round-precision=1}
\centering
\tiny
\caption{{DSC and ECE comparison of RobustSeg and mmFormer on BraTS 2020 and FeTS 2024.} Values are reported as ECE/DSC and averaged over modality combinations.}
\begin{tabularx}{\textwidth}{
    c | c | c | 
    >{\centering\arraybackslash\hsize=1.15\hsize}X 
    >{\centering\arraybackslash\hsize=1.15\hsize}X 
    >{\centering\arraybackslash\hsize=1.15\hsize}X 
    >{\centering\arraybackslash\hsize=1.15\hsize}X 
    >{\centering\arraybackslash\hsize=0.7\hsize}X 
    >{\centering\arraybackslash\hsize=0.7\hsize}X 
    >{\centering\arraybackslash\hsize=0.7\hsize}X 
    >{\centering\arraybackslash\hsize=0.7\hsize}X 
    >{\centering\arraybackslash\hsize=1.6\hsize}X
}
\toprule
Dataset & Model & Type 
& Uncal
& mL1-ACE 
& SDC 
& MC
& TS
& DirS
& LTS
& SelS
& MMA-LTS$^\ast$ \\
\midrule
\multirow{6}{*}{\shortstack{BraTS\\2020}}
& \multirow{3}{*}{RobustSeg} & WT & \num{20.68}/\num{84.09} & \num{19.68}/\num{80.13} & \num{20.10}/\num{83.53} & \num{20.73}/\num{84.31} & \num{19.54} & \num{20.47} & \num{15.57} & \num{15.92} & \cellcolor[HTML]{D6EAF8}\textbf{\num{14.93}} \\
& & \cellcolor[HTML]{EFEFEF}{TC} & \cellcolor[HTML]{EFEFEF}{\num{26.18}/\num{75.30}} & \cellcolor[HTML]{EFEFEF}{\num{23.96}/\num{71.30}} & \cellcolor[HTML]{EFEFEF}{\num{23.73}/\num{74.83}} & \cellcolor[HTML]{EFEFEF}{\num{25.50}/\num{75.44}} & \cellcolor[HTML]{EFEFEF}{\num{24.90}} & \cellcolor[HTML]{EFEFEF}{\num{23.96}} & \cellcolor[HTML]{EFEFEF}{\num{18.89}} & \cellcolor[HTML]{EFEFEF}{\num{18.63}} & \cellcolor[HTML]{D6EAF8}\textbf{\num{16.56}} \\
& & ET & \num{21.18}/\num{62.96} & \num{19.87}/\num{59.79} & \num{19.28}/\num{62.79} & \num{20.76}/\num{63.14} & \num{20.22} & \num{20.14} & \num{21.02} & \num{20.03} & \cellcolor[HTML]{D6EAF8}\textbf{\num{15.78}} \\
\cmidrule{2-12}
& \multirow{3}{*}{mmFormer} & WT & \num{22.10}/\num{86.32} & \num{20.30}/\num{86.06} & \num{20.37}/\num{86.62} & \num{22.17}/\num{86.13} & \num{21.49} & \num{22.05} & \num{17.01} & \num{18.40} & \cellcolor[HTML]{D6EAF8}\textbf{\num{11.61}} \\
& & \cellcolor[HTML]{EFEFEF}{TC} &  \cellcolor[HTML]{EFEFEF}{\num{29.11}/\num{79.70}} &  \cellcolor[HTML]{EFEFEF}{\num{25.27}/\num{78.65}} &  \cellcolor[HTML]{EFEFEF}{\num{26.32}/\num{79.65}} &  \cellcolor[HTML]{EFEFEF}{\num{29.15}/\num{79.56}} &  \cellcolor[HTML]{EFEFEF}{\num{28.43}} &  \cellcolor[HTML]{EFEFEF}{\num{28.43}} &  \cellcolor[HTML]{EFEFEF}{\num{19.38}} &  \cellcolor[HTML]{EFEFEF}{\num{24.88}} & \cellcolor[HTML]{D6EAF8}\textbf{\num{17.42}} \\
& & ET & \num{26.50}/\num{65.35} & \num{22.59}/\num{64.15} & \num{23.90}/\num{65.30} & \num{26.60}/\num{65.20} & \num{25.83} & \num{26.31} & \num{17.78} & \num{22.05} & \cellcolor[HTML]{D6EAF8}\textbf{\num{16.46}} \\
\midrule
\multirow{6}{*}{\shortstack{FeTS\\2024}}
& \multirow{3}{*}{RobustSeg} & WT & \num{15.93}/\num{86.66} & \num{17.63}/\num{85.28} & \num{15.01}/\num{86.91} & \num{16.50}/\num{86.91} & \num{13.92} & \num{15.19} & \num{11.42} & \num{10.90} & \cellcolor[HTML]{D6EAF8}\textbf{\num{9.45}} \\
& & \cellcolor[HTML]{EFEFEF}{TC} & \cellcolor[HTML]{EFEFEF}{\num{20.84}/\num{81.62}} & \cellcolor[HTML]{EFEFEF}{\num{19.36}/\num{79.45}} & \cellcolor[HTML]{EFEFEF}{\num{17.58}/\num{81.67}} & \cellcolor[HTML]{EFEFEF}{\num{20.88}/\num{81.68}} & \cellcolor[HTML]{EFEFEF}{\num{18.50}} & \cellcolor[HTML]{EFEFEF}{\num{18.50}} & \cellcolor[HTML]{EFEFEF}{\num{14.26}} & \cellcolor[HTML]{EFEFEF}{\num{14.86}} & \cellcolor[HTML]{D6EAF8}\textbf{\num{12.89}} \\
& & ET & \num{16.97}/\num{71.73} & \num{16.63}/\num{69.79} & \num{14.69}/\num{71.17} & \num{17.09}/\num{71.79} & \num{14.93} & \num{15.22} & \num{12.77} & \num{13.19} & \cellcolor[HTML]{D6EAF8}\textbf{\num{12.03}} \\
\cmidrule{2-12}
& \multirow{3}{*}{mmFormer} & WT & \num{16.63}/\num{86.75} & \num{14.50}/\num{87.36} & \num{11.88}/\num{86.74} & \num{16.81}/\num{86.46} & \num{15.33} & \num{16.66} & \num{12.75} & \num{16.34} & \cellcolor[HTML]{D6EAF8}\textbf{\num{11.83}} \\
& & \cellcolor[HTML]{EFEFEF}{TC} & \cellcolor[HTML]{EFEFEF}{\num{22.89}/\num{83.25}} & \cellcolor[HTML]{EFEFEF}{\num{15.57}/\num{82.79}} & \cellcolor[HTML]{EFEFEF}{\num{15.47}/\num{83.64}} & \cellcolor[HTML]{EFEFEF}{\num{22.97}/\num{83.14}} & \cellcolor[HTML]{EFEFEF}{\num{21.40}} & \cellcolor[HTML]{EFEFEF}{\num{22.48}} & \cellcolor[HTML]{EFEFEF}{\num{16.89}} & \cellcolor[HTML]{EFEFEF}{\num{22.29}} & \cellcolor[HTML]{D6EAF8}\textbf{\num{13.29}} \\
& & ET & \num{19.54}/\num{72.69} & \num{13.77}/\num{72.32} & \num{13.97}/\num{72.63} & \num{19.64}/\num{72.58} & \num{28.03} & \num{21.17} & \num{17.23} & \num{19.02} & \cellcolor[HTML]{D6EAF8}\textbf{\num{13.19}} \\
\bottomrule
\end{tabularx}
\label{tab:tab3}
\end{table}

\subsection {Experiment Results}
\begin{figure}[t]
    \centering
    \includegraphics[width=0.94\linewidth]{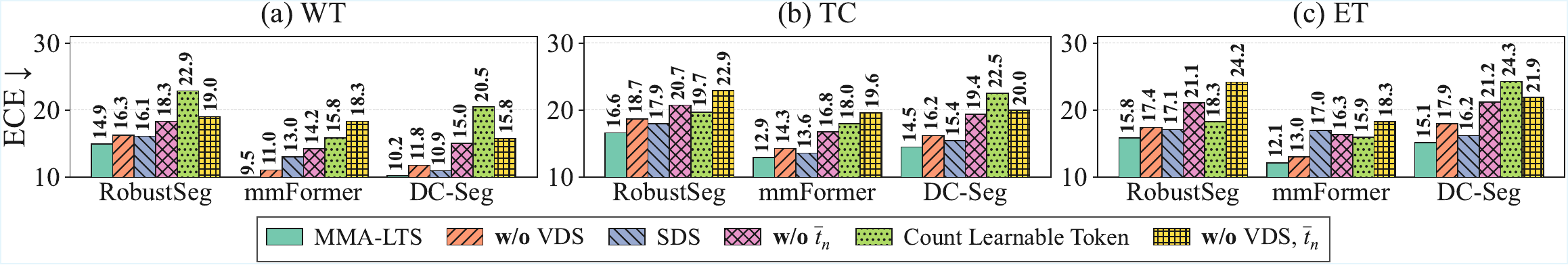}
    \caption{{Ablation study of MMA-LTS.} ECE are averaged over modality combinations.}
    \label{fig:figure3}
\end{figure}

\textbf{Comparative Experiments.} Tables \ref{tab:tab1} and \ref{tab:tab2} present results on BraTS 2020 and FeTS 2024. MMA-LTS achieves the lowest ECE across most tumor regions and 15 modality combinations while preserving DSC. On BraTS 2020, MMA-LTS reduces average WT/TC/ET ECE to 10.2/14.5/15.1, outperforming the second-best LTS (13.1/16.8/16.5). On FeTS 2024, MMA-LTS similarly achieves the best ECE across tumor regions (9.5/12.0/11.6), demonstrating effective correction of modality-combination-induced and spatially heterogeneous confidence distortions. Table~\ref{tab:tab3} further confirms consistent ECE reduction across RobustSeg and mmFormer backbones, validating its generalized calibration effectiveness.
\\
\textbf{Ablation Study.} Fig.~\ref{fig:figure3} presents ablation results of MMA-LTS on DC-Seg with BraTS 2020. Removing VDS or replacing it with sample-averaged difficulty ($\mathrm{SDS}$) consistently increases ECE across all tumor regions, highlighting the importance of spatial difficulty estimation. Removing the modality availability token $\bar{t}_n$ or substituting it with a count-based token also degrades ECE, as it is limited in capturing modality combination-specific calibration errors. These results demonstrate that each component contributes to calibration improvement across all tumor regions under missing modalities.

\begin{figure}[t!]
    \centering
    \includegraphics[width=0.98\linewidth]{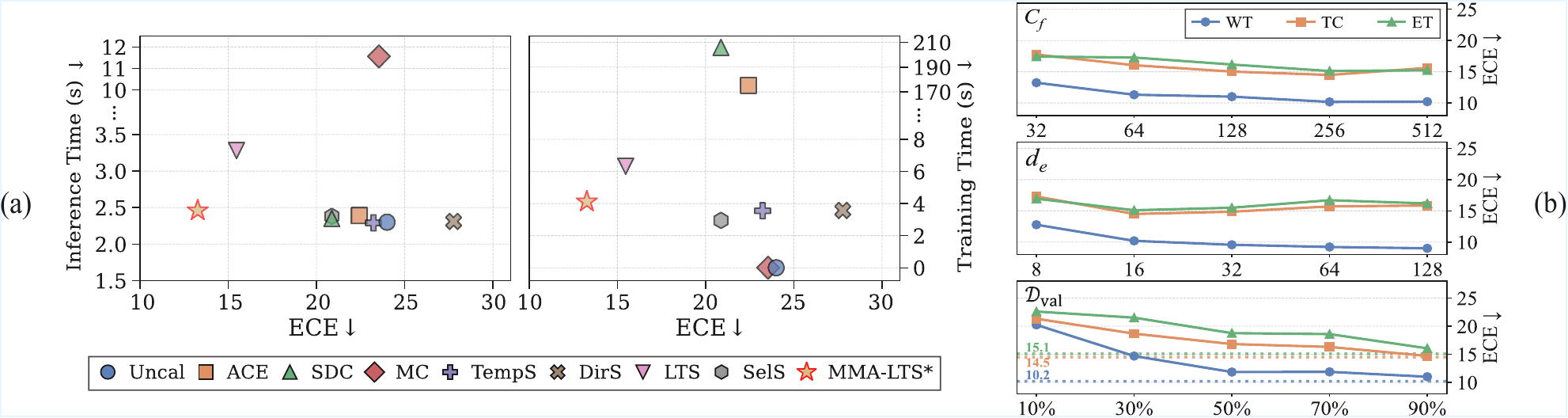}
    \caption{{Time efficiency and sensitivity analysis (DC-Seg, BraTS 2020).} (a) ECE versus inference and training time. (b) Sensitivity to $C_f$, $d_e$, and $\mathcal{D}_{\text{val}}$ size (dashed: 100\%).}
    \label{fig:figure4}
\end{figure}

\textbf{Qualitative Analysis.} 
Fig.~\ref{fig:figure1}(b) presents two cases with identical modality counts but different modality combinations (T2+FLAIR and T2+T1ce). Although their DSC values are comparable, the uncalibrated model exhibits different confidence patterns, with pronounced overconfident errors near tumor boundaries, assigning high tumor probability to non-tumor regions (false positives) and confidently mispredicting true tumors as background (false negatives), particularly in the T2+T1ce case. The VDS maps consistently highlight these ambiguous regions, while the calibration strength maps indicate where stronger correction is required for different modality combinations. MMA-LTS suppresses overconfident predictions and corrects spatially heterogeneous confidence distortions, yielding more reliable confidence estimates. Nonetheless, residual overconfidence near tumor boundaries remains, leaving room for improvement.
\\
\textbf{Time Efficiency and Sensitivity Analysis.} As a lightweight post-hoc module with a frozen backbone, MMA-LTS achieves the lowest ECE with inference time comparable to temperature scaling-based methods and much lower training cost than retraining approaches (mL1-ACE, SDC) (Fig.~\ref{fig:figure4}(a)). MC requires multiple forward passes and thus higher inference cost. Sensitivity analysis (Fig.~\ref{fig:figure4}(b)) confirms robustness to hyperparameters and $\mathcal{D}_{\text{val}}$ size, showing stable ECE for $C_f \geq 128$ and $d_e \geq 16$, with performance saturating beyond 50\% of $\mathcal{D}_{\text{val}}$.
\\
\section{Conclusion}
We present MMA-LTS, a plug-in post-hoc calibration method that corrects modality-combination-specific and spatially heterogeneous miscalibration in brain tumor segmentation under missing modalities. By conditioning on modality availability and voxel-wise difficulty, MMA-LTS enables spatially adaptive calibration without retraining the segmentation model or compromising accuracy. Experiments on BraTS 2020 and FeTS 2024 demonstrate consistent ECE reductions across multiple backbones and diverse missing-modality scenarios, establishing MMA-LTS as a practical step toward trustworthy clinical deployment. Future work includes extending validation to other anatomical domains and imaging modalities, incorporating reliability metrics beyond ECE with formal statistical testing, and further clinical validation in real-world settings.

\begin{credits}
\subsubsection{\ackname} This work was supported by the National Research Foundation of Korea (NRF) grant funded by the Korean government (MSIT) under grant number RS-2022-NR068758. We are also deeply grateful for the generous support provided by the Seegene Medical Foundation in South Korea.
\subsubsection{\discintname} The authors have no competing interests to declare that are relevant to the content of this article.
\end{credits}

\bibliographystyle{splncs04}
\bibliography{Paper-1236}

\end{document}